\documentclass[10pt,twocolumn,letterpaper]{article}

\usepackage{iccv}
\usepackage{times}
\usepackage{epsfig}
\usepackage{graphicx}
\usepackage{amsmath}
\usepackage{amssymb}

\usepackage{subfigure}
\usepackage{multirow}

\usepackage{verbatim}
\usepackage[pagebackref=true,breaklinks=true,letterpaper=true,colorlinks,bookmarks=false]{hyperref}

\usepackage{color}
\usepackage{cite}
\usepackage{url}

\usepackage{algorithm}
\usepackage{algorithmic}

\usepackage{caption}
\usepackage{verbatim}

\def\ie{\emph{i.e.~}}
\def\eg{\emph{e.g.~}}
\def\etal{{\em et al.~}}

\iccvfinalcopy 

\def\iccvPaperID{****} 

\begin{document}

\title{Adversarial Attacks against Deep Saliency Models}

\author{Zhaohui Che\\
Shanghai Jiao Tong University\\
\and
Ali Borji\\
MARKABLE\\
\and
Guangtao Zhai\\
Shanghai Jiao Tong University\\
\and
Suiyi Ling\\
Universit{\'e} de Nantes\\
\and
Guodong Guo\\
West Virginia University,\\
Baidu Research \\
\and
Patrick Le Callet\\
Universit{\'e} de Nantes\\
}

\maketitle

\begin{abstract}
Currently, a plethora of saliency models based on deep neural networks have led great breakthroughs in many complex high-level vision tasks (\eg scene description, object detection). The robustness of these models, however, has not yet been studied. In this paper, we propose a sparse feature-space adversarial attack method against deep saliency models for the first time. The proposed attack only requires a part of the model information, and is able to generate a sparser and more insidious adversarial perturbation, compared to traditional image-space attacks. These adversarial perturbations are so subtle that a human observer cannot notice their presences, but the model outputs will be revolutionized. This phenomenon raises security threats to deep saliency models in practical applications.
We also explore some intriguing properties of the feature-space attack, \eg~1) the hidden layers with bigger receptive fields generate sparser perturbations, 2) the deeper hidden layers achieve higher attack success rates, and 3) different loss functions and different attacked layers will result in diverse perturbations. Experiments indicate that the proposed method is able to successfully attack different model architectures across various image scenes.

\end{abstract}

\begin{figure}
\centering
\subfigure[Guide Image]{\label{fig:edge-a}\includegraphics[height=0.24\linewidth]{./img_guide.jpg}}
\vspace{-5pt}
\subfigure[Original Image]{\label{fig:edge-a}\includegraphics[height=0.24\linewidth]{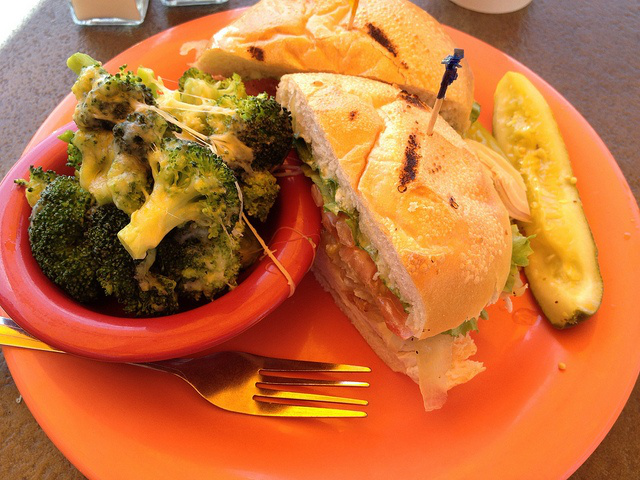}}
\subfigure[Original Output]{\label{fig:edge-a}\includegraphics[height=0.24\linewidth]{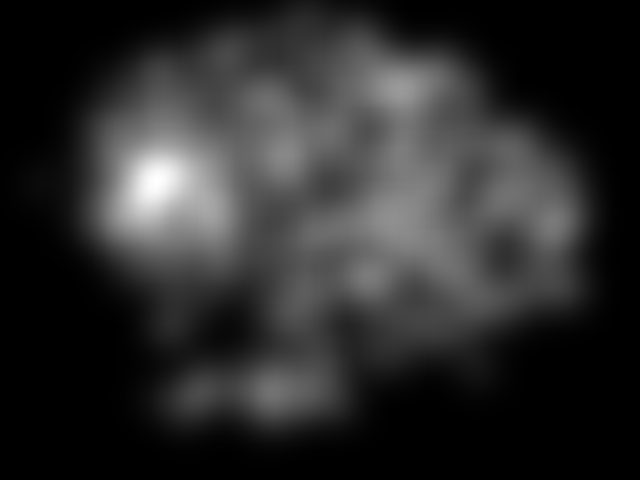}} \\

\subfigure[Image-space Perturbation]{\label{fig:edge-a}\includegraphics[height=0.24\linewidth]{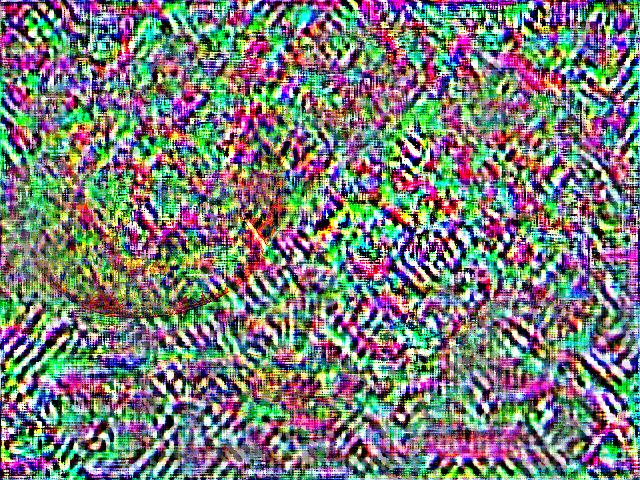}}
\subfigure[Image-space Adv-ersarial Example]{\label{fig:edge-a}\includegraphics[height=0.24\linewidth]{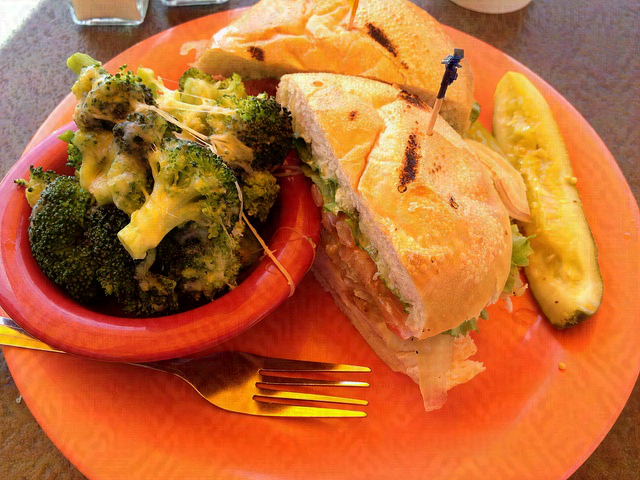}} 
\subfigure[Output of Image-space Adv. Example]{\label{fig:edge-a}\includegraphics[height=0.24\linewidth]{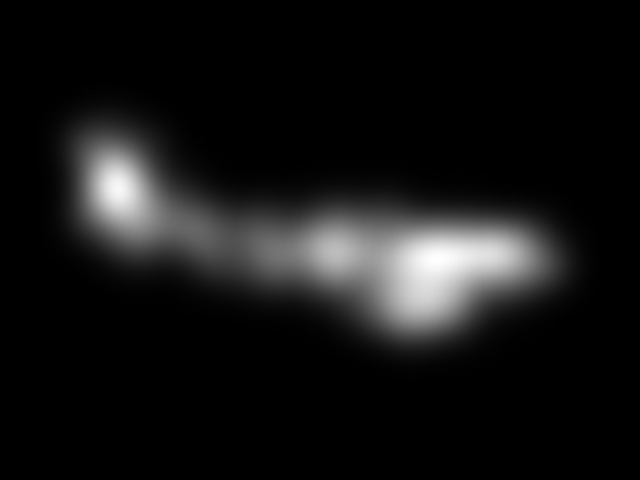}}\\

\vspace{-5pt}
\subfigure[Feature-space Perturbation]{\label{fig:edge-a}\includegraphics[height=0.24\linewidth]{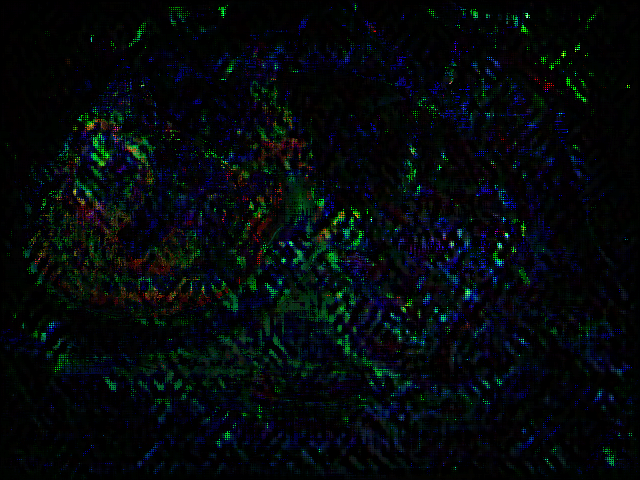}}
\subfigure[Feature-space Adv-ersarial Example]{\label{fig:edge-a}\includegraphics[height=0.24\linewidth]{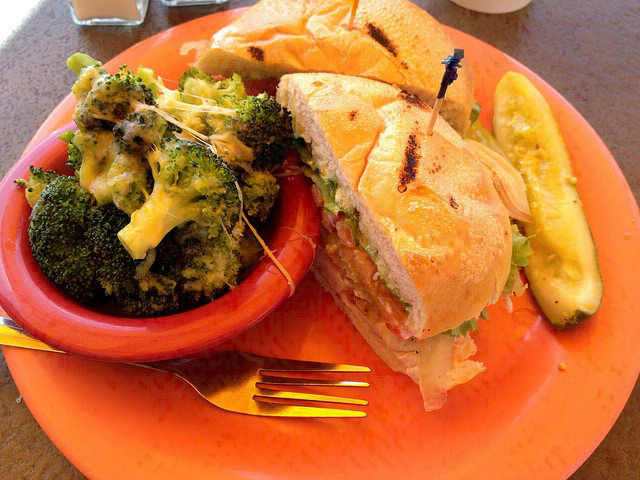}} 
\subfigure[Output of Feature-space Adv. Example]{\label{fig:edge-a}\includegraphics[height=0.24\linewidth]{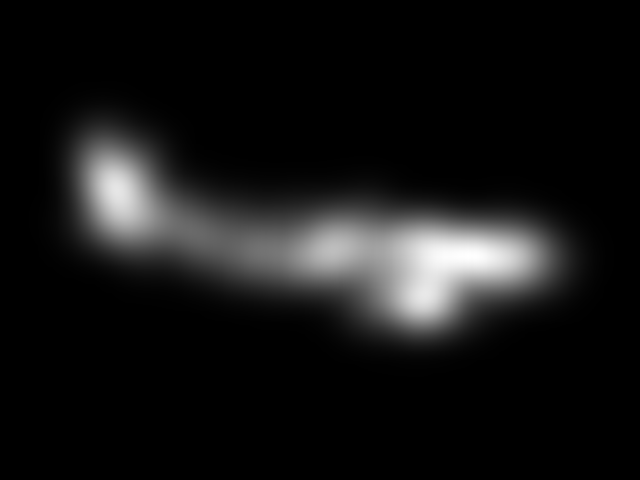}}\\

\caption{\small{The targeted adversarial examples generated by the proposed feature-space and image-space attacks. Both image-space and feature-space attacks are able to change the model output toward the guide image by adding some negligible perturbations to the original image.
Besides, as shown in (d) and (g), the adversarial perturbation generated by the feature-space attack is much sparser than that of the image-space attack. We suggest the readers to zoom in the figures, especially (b), (e) and (h), for more details and better observation.
Feature-space adversarial attack is more visually imperceptible and is more dangerous to security-related applications \cite{smautodrive,dataattmotodrive}. Notably, the adversarial perturbations in (d) and (g) are normalized by the $min$-$max$ linear normalization method for better observation. In fact, the adversarial perturbations are visually negligible because the maximum intensity in each channel is less than 0.07 (the normalized image intensity values are ranging from 0 to 1). SalGAN \cite{SalGAN} serves as the threat model here.}}
\label{beginfigure}
\vspace{-15pt}
\end{figure}

\begin{figure}
\centering
\subfigure[Nontargeted Adv. Example by KL loss]{\label{fig:edge-a}\includegraphics[height=0.24\linewidth]{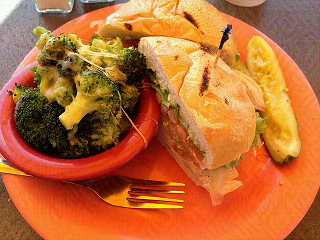}}
\subfigure[Nontargeted Adv. Example by CC loss]{\label{fig:edge-a}\includegraphics[height=0.24\linewidth]{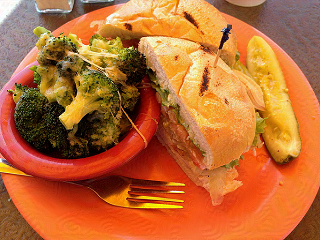}} 
\subfigure[Nontargeted Adv. Example by L$_1$ loss]{\label{fig:edge-a}\includegraphics[height=0.24\linewidth]{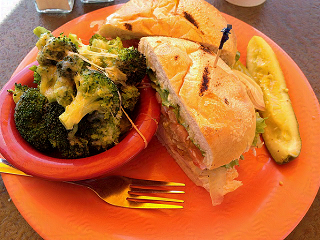}}\\

\subfigure[Output of Adv. Example by KL loss]{\label{fig:edge-a}\includegraphics[height=0.24\linewidth]{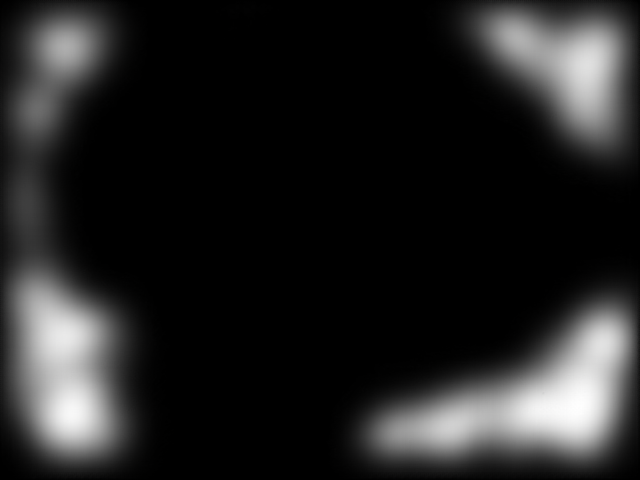}}
\subfigure[Output of Adv. Example by CC loss]{\label{fig:edge-a}\includegraphics[height=0.24\linewidth]{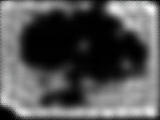}} 
\subfigure[Output of Adv. Example by L$_1$ loss]{\label{fig:edge-a}\includegraphics[height=0.24\linewidth]{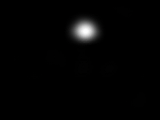}}

\caption{\small The nontargeted adversarial examples generated by different losses. We notice that the non-targeted adversarial perturbations are highly depended on loss functions, and final saliency maps of different perturbations are totally different. SALICON \cite{SALICON} serves as the threat model here.}
\label{beginfigure2}
\end{figure}

\section{Introduction}

Human visual attention is an advanced internal mechanism for selecting informative and conspicuous regions from external visual stimuli. A plethora of saliency models based on deep neural networks \cite{MLnet,SalGAN,SALICON,SalNet,SAM} have been proposed in the past decades to predict human gaze by simulating biological attention mechanisms~\cite{borji2013state,borji2013quantitative,Borji_2013_ICCV}. Bottom-up saliency is an efficient front-end process to complex back-end high-level vision tasks such as scene understanding, finer-grained classification, object recognition, visual description, and security-related automatic drive applications \cite{SalObjDec01,SalObjDec02,SalSegment03,smautodrive,dataattmotodrive}. 

However, most of current deep neural networks are highly vulnerable to the adversarial examples \cite{nipscompe}, which are generated by adding some negligible but deliberate adversarial perturbations to the original image, as shown in Fig.\ref{beginfigure}. The adversarial perturbation is so subtle that a human observer cannot notice its presence, but the model outputs will be revolutionized.  This phenomenon raises security threats to deep networks in practical applications.

In this paper, we dig into the feature space of deep networks, and investigate the feature-space attack against saliency detection task. We ask and answer the following questions: $\bf(1)$ Which layers can perform successful feature-space attack? $\bf(2)$ What is the difference between perturbations generated by image-space and feature-space attacks? $\bf(3)$ Where (and at which layer) does the output of the threat model start to spoil when we feed the adversarial example?, and $\bf(4)$ How can we defend these attacks? To the best of our knowledge, this is the first work addressing adversarial attacks against saliency models.

\subsection{Related works and concepts}

Most of previous works mainly focused on image classification tasks \cite{fgsm,bfgs,itefgsm,CandW,ATNnetworl}, which aimed to fool the image classifiers to predict a wrong category label. Recently, Xie~\etal~\cite{iccvsegadv} extended the adversarial attack to semantic segmentation and object detection fields using a dense adversary generation method. Generating the adversarial examples for semantic segmentation and object detection tasks are much more difficult than classification task, because there are more threat targets to be attacked, \eg~multiple pixels or proposals. Metzen~\etal~\cite{iccvuniper} explored the universal adversarial perturbations for semantic segmentation task, and verified the existence of universal perturbations for segmentation models. Universal perturbations are input-agnostic, and are able to fool the deep networks on the majority of various input images. Zeng~\etal~\cite{cvpr3d2d} investigated the subset of adversarial examples that correspond to meaningful changes in 3D physical properties (\ie~rotation, translation, and illumination conditions), and pointed out that the adversarial attack in 3D physical space is more difficult than traditional 2D image space.

One explanation for adversarial example is that the subtle adversarial perturbation falls on some areas in the large and high-dimensional feature space which has not been explored in the training stage \cite{iccvsegadv}.
Compared to image classification task, saliency detection is a basic pixel-to-pixel translation problem, which provides an opportunity to directly observe the internal attention regions of deep networks across different levels of representations. Therefore, investigating the feature-space adversarial attack against the saliency models helps to understand the internal attention mechanism of deep networks, and may provide the potential possibilities to promote deep networks' robustness.

Most of previous works perform the adversarial attacks from image-space. For semantic segmentation task \cite{iccvsegadv,iccvuniper}, the image-space attack generates the adversarial perturbation by calculating the loss between the predicted segmentation labels of the original image and the guide image in image space (\ie~the predicted segmentation results have the shape of height$\times$width$\times$channels, normally channels=3 for RGB image), then back-propagates the gradients of image-space loss with respect to the original image through the entire threat model to generate the adversarial perturbation. Notably, the threat model means the pretrained model to be attacked. The adversarial attacks may have knowledge of the threat model including parameters and architectures, but are not allowed to modify the threat model \cite{tnnreview}.

Accordingly, the feature-space attack generates adversarial perturbation by computing the loss between the intermediate representations (\ie~the shape of the intermediate representation is height$\downarrow$$\times$width$\downarrow$$\times$channels$\uparrow$, where height$\downarrow$ and width$\downarrow$ represent the downsampled image size, while channels$\uparrow$ represents the extended channels, \eg~channels$\uparrow$=1024 in the $4_{th}$ convolutional layer of ResNet-101 \cite{resnetpaper}) between original image and guide image, then back-propagates the gradients through the layers before the attacked layer, rather than the entire network.
Thus, the feature-space attack requires less information about the threat model, and manipulates the intermediate high-dimensional representations with smaller spatial resolutions but greater channels/dimensions.

Generally speaking, most of the adversarial attacks can be categorized along different dimensions. First, attacks can be classified by the type of desired output. In $\textup{\bf Targeted Attack Scenario}$, the attacks aim to change the threat model's output toward some specific guide direction. In $\textup{\bf Nontargeted Attack Scenario}$, the attacks only need to destroy the original correct prediction, but the specific guide direction does not matter.
Second, attacks can also be classified by the amount of knowledge that the attacks have already known about the threat model \cite{nipscompe}. $\textup{\bf White Box Attacks}$ have full knowledge of the entire threat model. $\textup{\bf Black Box Attacks}$ have no knowledge about the threat model. $\textup{\bf Black Box Attacks with Probing}$ do not have full information about the threat model, but can obtain a part of parameters and architectures of the threat model. In this paper, the proposed feature-space attack is a variant of the Black Box Attack with Probing, while the image-space attack is a White Box attack.

A previous work studied adversarial examples in the feature space for image classification task. Sabour~\etal~\cite{featureiclr} tried to fool the image classifier by minimizing the distance of the internal representations of the original image and the guide image by a non-iterative optimization method. However, this work has not revealed the sparsity of feature-space adversarial attack, and has not investigated the finer-grained relations between feature-space perturbation and its diversity, perceptibility and aggressivity.

\subsection{Our contributions}
Our contributions and lessons include:
\begin{itemize}
    \item {\bf } We propose the targeted and nontargeted adversarial attack methods against deep saliency models. The proposed methods are able to perform successful attacks in both image-space and feature-space across different network architectures and various image scenes.
    \item {\bf } The proposed feature-space attack only requires a part of information of the threat model, and can completely change the output of entire model. The generated perturbation is sparser and more imperceptible compared to traditional image-space attack, as shown in Fig.\ref{beginfigure}.
    \item {\bf } On the one hand, the success rate of feature-space attack is highly related to the depth of the attacked layer, \ie~the deeper layers which detect more semantic information achieve the higher success rates, while the shallower layers which extract tiny texture and edges always fail to generate an effective adversarial perturbation. On the other hand, the perceptibility/sparsity of the generated perturbation is highly depended on the receptive field of the attacked layer, \eg~when we attack the SalGAN \cite{SalGAN} model which utilizes the classic encoder-decoder architecture, the context hidden layer
        between the encoder layers and the decoder layers, which has the biggest receptive field, achieves the sparsest adversarial perturbation compared to other layers and image-space attack.
    \item {\bf } For targeted attack scenario, when we fix the loss function, the perturbations from different attacked layers have discrepant patterns, but result in similar model predictions toward guide image. Besides, when we fix the position of attacked layer, the perturbations generated by different losses also result in similar outputs.
    \item {\bf } For the nontargeted attack scenario, different attacked layers and different loss functions will produce diverse adversarial perturbations, and result in totally different outputs,  as shown in Fig.\ref{beginfigure2}.
    \item {\bf } For defending nontargeted attack, we find that the perturbations generated by image-space attack are able to mitigate the perturbations generated by feature-space attack to a certain extent, however, the perturbations generated by feature-space attack can not countervail the image-space attack.
        The targeted attack is more difficult to defend, because perturbations generated by different losses cannot countervail with each other. 
\end{itemize}


\section{Proposed Adversarial Attack Method}

\begin{figure*}
\center
\includegraphics[scale=0.26]{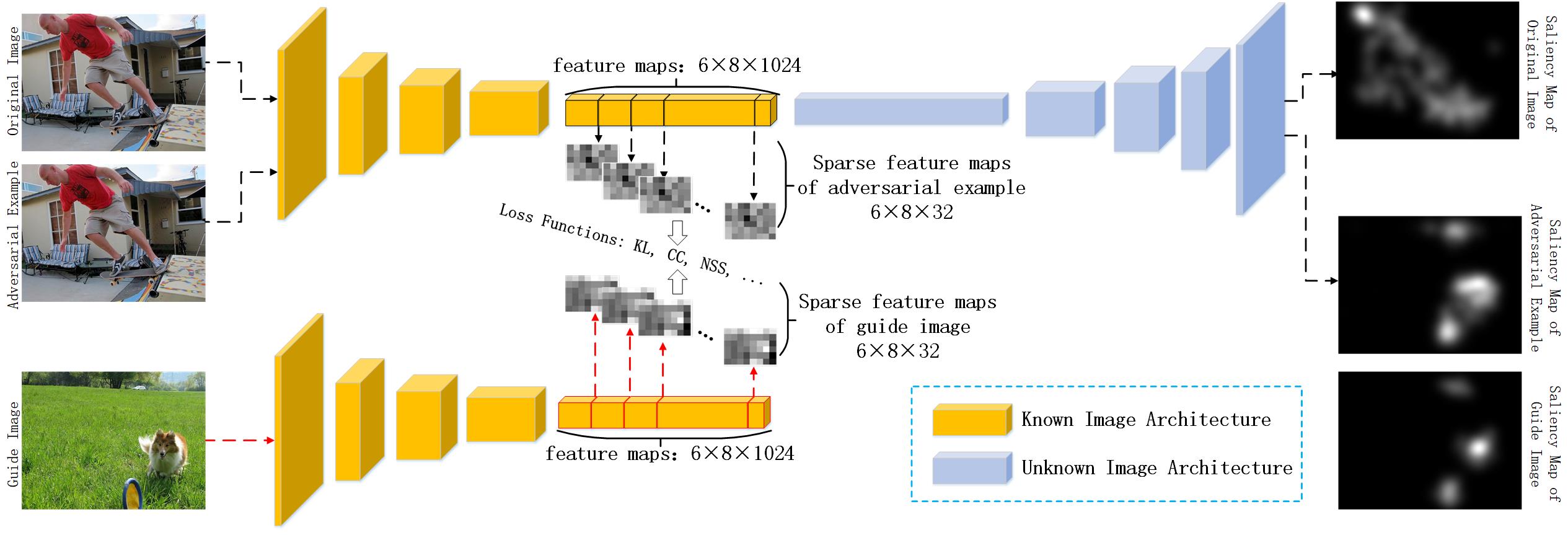}
\caption{\small This diagram illustrates the main idea behind the proposed sparse feature-space adversarial attack method. In this figure, we select the SalGAN model \cite{SalGAN}, which utilizes a classic encoder-decoder architecture, as the threat model. Specifically, we perform the attack from the context hidden layer between encoder layers and decoder layers of SalGAN model, which obtains 1024 feature maps with small scale resolution of 6$\times$8. This is because the context hidden layer has the biggest receptive field, and encodes the most important semantic information into the high-dimensional representations.
We further uniformly select 32 feature maps from the 1024 feature maps of the adversarial example and the guide image respectively. Notably, the positions of the selected sparse feature maps from adversarial example and guide image are the same. Then we compute the losses between the selected sparse feature map pairs, and back-propagate the gradients of the feature-space loss with respect to the adversarial example through the known layers before the attacked layer. This way, we generate a subtle adversarial perturbation, which thoroughly changes the final prediction of the entire threat model. This method can be extended to attack any layer. When we attack the final output layer, this method becomes the image-space attack. The sparsity of the proposed method has two meanings. First, we select a fraction of feature maps from hundreds of feature maps of the attacked layer to perform the attack. Second, the generated adversarial perturbation is sparse and visually imperceptible.}
\label{flowchart}
\end{figure*}

\subsection{Targeted Attack Method}

Unlike traditional attacks for image classification \cite{fgsm,bfgs,itefgsm} and semantic segmentation \cite{iccvsegadv} tasks, which aim to change the predicted category label of the entire image or pixels/proposals, we perform the attack by reducing the distance between the high-dimensional representations of the adversarial example and the guide image. The general idea behind the proposed method is shown in Fig.\ref{flowchart}.

To generate the targeted adversarial perturbation against the saliency detection model, the goal is distracting the model attention from the salient regions of the original image towards the salient regions of the guide image, at the same time, keeping that the adversarial example looks almost the same as the original image, \ie~:

\begin{equation}
\small
\left\{
             \begin{array}{lr}
             \textup{arg min}~~~~\mathbb{D}_0[\mathbb{F}^{\downarrow}_i({\bf G}), ~~\mathbb{F}^{\downarrow}_i({\bar{\bf I} })],  \\
             \textit{in order to:} ~~~\textup{arg min}~\mathbb{D}_1[\mathbb{F}(\textbf{G}), ~~\mathbb{F}(\bar{\bf I})],  \\
             \textit{subject to:} ~~~\mathbb{D}_2[\textup{\bf I}, ~~\bar{\bf I}] \approx 0. \\
             \end{array}
\right.
\label{eq1}
\end{equation}
where $\mathbb{F}$ means the threat model, while $\mathbb{F}(\textbf{G})$ and $\mathbb{F}(\bar{\bf I})$ represent the predicted saliency maps of the guide image $\textbf{G}$ and the adversarial example $\bar{\bf I}$. In addition, $\mathbb{F}^{\downarrow}_i({\bf G})$ and $\mathbb{F}^{\downarrow}_i({\bar{\bf I} })$ represent the high-dimensional representations of $\textbf{G}$ and $\bar{\bf I}$, which are selected from the $i_{th}$ convolution layer of $\mathbb{F}$.
Besides, $\mathbb{D}_0$ measures the distance between high-dimensional representations of the adversarial example and the guide image,
while $\mathbb{D}_1$ measures the distance between model predictions of the adversarial example and the guide image, and $\mathbb{D}_2$ means the perceptual similarity between original input $\textup{\bf I}$ and the adversarial example $\bar {\bf I}$. The smaller values of $\mathbb{D}_0$, $\mathbb{D}_1$ and $\mathbb{D}_2$ mean the higher similarity.

In most cases, we cannot obtain the full information about the threat model. Suppose that we can only obtain the information of some front-end layers of the threat model, we can perform the attack from a known convolution layer as shown in Fig.\ref{flowchart}. Specifically, we feed the adversarial example $\bar{\bf I}$ into the threat model $\mathbb{F}$, and obtain the $6\times8\times1024$ feature maps $\mathbb{F}_i(\bar{\bf I})$ from the $i_{th}$ hidden layer $\mathbb{F}_i$. Next, we select a fraction of feature maps $\mathbb{F}^{\downarrow}_i(\bar{\bf I})$ from $\mathbb{F}_i(\bar{\bf I})$, \eg~we uniformly select 32 feature maps from the 1024 feature maps of the context hidden layer of SalGAN model, as shown in Fig.\ref{flowchart}. In Section-3.3, we further discuss how many sparse feature maps should we select to guarantee a successful attack. Similarly, we select 32 sparse feature maps $\mathbb{F}^{\downarrow}_i(\textup{\bf G})$ of guide image $\textup{\bf G}$. Importantly, the position of each $6\times8\times1$ feature map in $\mathbb{F}^{\downarrow}_i(\textup{\bf G})$ is the same as that of $\mathbb{F}^{\downarrow}_i(\bar{\bf I})$, because we tried to permute and mismatch the feature maps from $\mathbb{F}^{\downarrow}_i(\textup{\bf G})$ and $\mathbb{F}^{\downarrow}_i(\bar{\bf I})$, but failed to attack the model.
Then, we calculate the feature-space loss between $\mathbb{F}^{\downarrow}_i(\bar{\bf I})$ and $\mathbb{F}^{\downarrow}_i(\textup{\bf G})$:
\begin{equation}
\small
\mathbb{L}[\mathbb{F}^{\downarrow}_i(\bar{\bf I}), \mathbb{F}^{\downarrow}_i(\textup{\bf G})] = \frac{1}{N} ~\sum_{n=1}^{N} \textup{KL} [\mathbb{F}^{\downarrow}_{i_n}(\bar{\bf I}), \mathbb{F}^{\downarrow}_{i_n}(\textup{\bf G})],
\label{eq2}
\end{equation}
where $N$=32 is the amount of channels of selected sparse feature maps, and $\mathbb{F}^{\downarrow}_{i_n}(\bar{\bf I})$ is the $n_{th}$ $6\times8\times1$ feature map within $\mathbb{F}^{\downarrow}_{i}(\bar{\bf I})$.
$\textup{KL}$ represents the Kullback-Leibler Divergence metric \cite{mit-saliency-benchmark} which is wildly used to measure the similarity between two saliency maps, and the smaller $\textup{KL}$ score means the higher similarity. We can adopt some other loss functions to replace KL metric, \eg~Pearson's Linear Coefficient (CC) metric \cite{mit-saliency-benchmark}, Normalized Scanpath Saliency (NSS) metric \cite{nssdef}, and $\textup{L}_1$ distance. We further compare the performance of difference losses in Section-3.2.

By minimizing $\mathbb{L}[\mathbb{F}^{\downarrow}_i(\bar{\bf I}), \mathbb{F}^{\downarrow}_i(\textup{\bf G})]$, especially when attacking the layers which extract the high-level semantic information, we can reduce the distance between high-dimensional representations of the adversarial example and the guide image. This way, the final model prediction will be modified towards the salient regions of guide image. Surprisingly, by attacking some layers with big receptive fields, we can generate the much sparser perturbations compared to attacking final output layer, because the high-dimensional representations in these layers are able to distract model attention to the explicit target salient region, while keeping the smooth background region undisturbed. In other word, the final model prediction is highly depended on some intermediate hidden layers. If we can obtain the information of these layers, we can easily destroy the entire network. We further explore which hidden layers can generate sparse and valid perturbations in Section-3.3.

We use an iterative gradient descent optimization method to generate the targeted adversarial example. We set the initial perturbation ${\bf \Delta}_0=0$, and set the initial adversarial example $\bar{\bf I}_0=\bf I + {\bf \Delta}_0 = \bf I$. For the $t_{th}$ iteration, we obtain:
\begin{equation}
\small
\left\{
             \begin{array}{lr}
             {\bf \bar \Delta}_{t} = \frac{\partial \mathbb{L}[\mathbb{F}^{\downarrow}_i({\bar{\bf I}_t}), ~~\mathbb{F}^{\downarrow}_i({\bf G})]}{\partial \bar{\bf I}_t},  \\
             {\bf \Delta}_{t} = \gamma ~\times~ \frac{{\bf \bar \Delta}_{t} - min({\bf \bar \Delta}_{t})}{max({\bf \bar \Delta}_{t}) - min({\bf \bar \Delta}_{t}) + \epsilon}, \\
             \bar{\bf I}_{t+1} = \bar{\bf I}_{t} - \alpha{\bf \Delta}_{t},  \\
             \end{array}
\right.
\label{eq3}
\end{equation}
where $\alpha=2\times10^{-3}$ is the step size to control the magnitude of the gradient descent. And $\gamma=0.07$ is a scaling hyper-parameter to limit the intensity of perturbation. $\epsilon=1\times10^{-8}$ is a very small positive number, which is used to guarantee the divisor nonzero. The targeted attack algorithm terminates until that the model prediction of the adversarial example is almost the same as that of guide image, \ie~$\mathbb{D}_1[\mathbb{F}(\textbf{G}), ~~\mathbb{F}(\bar{\bf I}_{t})] < \tau_1$, where $\tau_1$ is the termination threshold, or it reaches the maximum iteration number.

\subsection{Nontargeted Attack Method}
In nontargeted attack scenario, the goal is distracting the model attention away from the salient regions of the original input, and there is no explicit guide image. We achieve this by maximizing the distance between high-dimensional representations of original input and the adversarial example, while keeping that the original and the adversarial example look the same. Similar to the targeted attack, we perform the nontargeted attack as follows:
\begin{equation}
\small
\left\{
             \begin{array}{lr}
             {\bf \bar \Delta}_{t} = \frac{\partial \mathbb{L}[\mathbb{F}^{\downarrow}_i({\bar{\bf I}_t}), ~~\mathbb{F}^{\downarrow}_i({\bf I})]}{\partial \bar{\bf I}_t},  \\
             {\bf \Delta}_{t} = \gamma ~\times~ \frac{{\bf \bar \Delta}_{t} - min({\bf \bar \Delta}_{t})}{max({\bf \bar \Delta}_{t}) - min({\bf \bar \Delta}_{t}) + \epsilon}, \\
             \bar{\bf I}_{t+1} = \bar{\bf I}_{t} + \alpha{\bf \Delta}_{t}.  \\
             \end{array}
\right.
\label{eq4}
\end{equation}

The nontargeted attack algorithm terminates until that the distance between the model predictions of adversarial example and original image are large enough, \ie~$\mathbb{D}_1[\mathbb{F}(\textbf{I}), ~~\mathbb{F}(\bar{\bf I}_{t})] > \tau_2$.

\begin{algorithm}[htb]
\footnotesize
\caption{: Sparse Feature-Space Adversarial Attack}
\label{alg:Framwork}
\begin{algorithmic}[1] 
\REQUIRE ~~\\ 
original image \textbf{I};
guide image \textbf{G};
the threat model $\mathbb{F}$; \\
the index $i$ of the intermediate layer of $\mathbb{F}$ to be attacked; \\
the amount of channels $N$ of selected feature maps; \\
the step size of iterative gradient descent $\alpha$; \\
the maximum iterations \textbf{X}; the saliency map evaluation metric $\mathbb{D}_1$;\\
the termination thresholds $\tau_1$ and $\tau_2$ for targeted and nontargeted attacks, respectively;
\ENSURE ~~\\ 
the adversarial example $\bar{\bf I}$; \\
\label{ code:fram:extract }
\STATE Initialization: ${\bf\Delta}_0 \leftarrow 0$; $\bar{\bf I}_0 \leftarrow {\bf I} + {\bf\Delta}_0$; $t \leftarrow 0$;
\label{code:fram:trainbase}
\IF{(Targeted Attack)}
\STATE Select $N$ sparse feature maps $\mathbb{F}^{\downarrow}_i(\textup{\bf G}))$ from $\mathbb{F}_i(\textup{\bf G})$;
\WHILE {($t < \textbf{X}$ ~~or~~ $\mathbb{D}_1[\mathbb{F}(\textbf{G}), ~~\mathbb{F}(\bar{\bf I}_{t})] \geq \tau_1$)}
\STATE Select $N$ feature maps $\mathbb{F}^{\downarrow}_i({\bar{\bf I}_t}))$ from $\mathbb{F}_i({\bar{\bf I}_t})$;
\STATE ${\bf \bar \Delta}_{t} \leftarrow \frac{\partial \mathbb{L}[\mathbb{F}^{\downarrow}_i({\bar{\bf I}_t}), ~~\mathbb{F}^{\downarrow}_i({\bf G})]}{\partial \bar{\bf I}_t}$;
\STATE ${\bf \Delta}_{t} \leftarrow \gamma \times \frac{{\bf \bar \Delta}_{t} - min({\bf \bar \Delta}_{t})}{max({\bf \bar \Delta}_{t}) - min({\bf \bar \Delta}_{t}) + \epsilon}$;
\STATE $\bar{\bf I}_{t+1} \leftarrow \bar{\bf I}_{t} - \alpha{\bf \Delta}_{t}$;
\STATE $t \leftarrow t + 1$;
\ENDWHILE
\ELSE
\STATE Select $N$ sparse feature maps $\mathbb{F}^{\downarrow}_i(\textup{\bf I}))$ from $\mathbb{F}_i(\textup{\bf I})$;
\WHILE {($t < \textbf{X}$ ~~or~~ $\mathbb{D}_1[\mathbb{F}(\textbf{I}), ~~\mathbb{F}(\bar{\bf I}_{t})] \leq \tau_2$)}
\STATE Select $N$  feature maps $\mathbb{F}^{\downarrow}_i({\bar{\bf I}_t}))$ from $\mathbb{F}_i({\bar{\bf I}_t})$;
\STATE ${\bf \bar \Delta}_{t} \leftarrow \frac{\partial \mathbb{L}[\mathbb{F}^{\downarrow}_i({\bar{\bf I}_t}), ~~\mathbb{F}^{\downarrow}_i({\bf I})]}{\partial \bar{\bf I}_t}$;
\STATE ${\bf \Delta}_{t} \leftarrow \gamma \times \frac{{\bf \bar \Delta}_{t} - min({\bf \bar \Delta}_{t})}{max({\bf \bar \Delta}_{t}) - min({\bf \bar \Delta}_{t}) + \epsilon}$;
\STATE $\bar{\bf I}_{t+1} \leftarrow \bar{\bf I}_{t} + \alpha{\bf \Delta}_{t}$;
\STATE $t \leftarrow t + 1$;
\ENDWHILE
\ENDIF
\RETURN $\bar{\bf I}$$\leftarrow$$\bar{\bf I}_{t}$. 
\end{algorithmic}
\end{algorithm}

\section{Experiments and Discussions}

\begin{table*}
\centering
\renewcommand{\arraystretch}{1.1}
\renewcommand{\tabcolsep}{2.3mm}

    \caption{\small Performance of the proposed feature-space attack method on different threat model architectures. For the $4_{th}$-$6_{th}$ columns, from left to right, the performance scores represent \textbf{CC, sAUC}, and \textbf{SIM} metrics \cite{Bylinskii} respectively, which measure the similarity between model predictions and human fixation ground truth. The higher CC, sAUC and SIM values mean the better performance.
    The $3_{rd}$ column represents the intermediate layer of the threat model to be attacked. The $4_{th}$, $5_{th}$ and $6_{th}$ columns represent the performance of threat model on original image, the adversarial examples generated by targeted  and nontargeted attacks, respectively.} 
    \vspace{4pt}
    \label{tab:model}
    \footnotesize
    \centering
    \begin{tabular}{|l|l|l|l|l|l|}

    \hline
    {\bfseries Threat Model} &{\bfseries Base Network} &{\bfseries Attacked Layer} &{\bfseries Original Performance } &{\bfseries Targeted Attack} &{\bfseries Nontargeted Attack }\\
   \hline
   \hline

   \hline
    {\bfseries{ \textup{SALICON$_1$}} \cite{SALICON}} &{{ multi-stream VGG-16}} &{\emph{ conv5-3-fine}} &{{0.748}, {0.726}, {0.723}} &{ {0.279}, {0.501}, {0.469}} &{ {-0.262}, {0.390}, {0.041}}\\
    {\bfseries{ \textup{SALICON$_2$}} \cite{SALICON}} &{{ multi-stream GoogleLenet}} &{\emph{inception4-fine}} &{{0.755}, {0.731}, {0.742}} &{ {0.288}, {0.514}, {0.457}} &{ {-0.255}, {0.385}, {0.042}}\\
    {\bfseries{ \textup{SALICON$_3$}} \cite{SALICON}} &{{ multi-stream AlexNet}} &{\emph{ conv5-1-fine}} &{{0.719}, {0.705}, {0.691}} &{ {0.291}, {0.520}, {0.463}} &{ {-0.280}, {0.382}, {0.039}}\\
    {\bfseries{ \textup{SALICON$_4$}} \cite{SALICON}} &{{ single-stream VGG-16}} &{\emph{ conv5-3-coarse}} &{{0.700}, {0.695}, {0.714}} &{ {0.266}, {0.482}, {0.428}} &{ {-0.307}, {0.346}, {0.035}}\\
    \hline
        \hline
    {\bfseries{ \textup{GazeGAN$_1$}} \cite{gazeganweb}} &{{ single-stream U-Net}} &{\emph{ decoder1-coarse}} &{ {0.796}, {0.740}, {0.764}} &{ {0.388}, {0.505}, {0.513}} &{ {-0.424}, {0.371}, {0.060}}\\
    {\bfseries{ \textup{GazeGAN$_2$}} \cite{gazeganweb}} &{{ multi-stream U-Net}} &{\emph{ decoder1-fine}} &{ {0.808}, {0.743}, {0.769}} &{ {0.419}, {0.511}, {0.534}} &{ {-0.356}, {0.382}, {0.069}}\\
    \hline
        \hline
    {\bfseries{ \textup{SalGAN}$_1$} \cite{SalGAN}} &{{ single-stream VGG-19}} &{\emph{  conv5-3-coarse}} &{ {0.703}, {0.707}, {0.713}} &{ {0.365}, {0.502}, {0.468}} &{ {-0.471}, {0.354}, {0.055}}\\
    {\bfseries{ \textup{SalGAN}$_2$} \cite{SalGAN}} &{{ multi-stream VGG-19}} &{\emph{ conv5-3-fine}} &{ {0.719}, {0.715}, {0.722}} &{ {0.373}, {0.511}, {0.486}} &{ {-0.454}, {0.360}, {0.058}}\\
    \hline
    \hline
    {\bfseries{ \textup{Global pix2pix}}\cite{pix2pixHD}} &{{ single-stream ResNet}} &{\emph{ res9-coarse}} &{ {0.766}, {0.729}, {0.748}} &{ {0.289}, {0.498}, {0.469}} &{ {-0.164}, {0.475}, {0.046}}\\
    {\bfseries{ \textup{Local pix2pix}}\cite{pix2pixHD}} &{{ multi-stream ResNet}} &{\emph{ res3-fine}} &{ {0.773}, {0.737}, {0.751}} &{ {0.313}, {0.504}, {0.435}} &{ {-0.188}, {0.490}, {0.048}}\\
    \hline
    \hline
    {\bfseries{ \textup{DVA}$_1$} \cite{dva}} &{{ single-stream VGG-16}} &{\emph{ conv5-3-coarse}} &{ {0.774}, {0.733}, {0.754}} &{ {0.308}, {0.514}, {0.517}} &{ {-0.400}, {0.415}, {0.054}}\\
    {\bfseries{ \textup{DVA}$_2$} \cite{dva}} &{{ multi-stream VGG-16}} &{\emph{ conv5-3-fine}} &{ {0.782}, {0.736}, {0.760}} &{ {0.312}, {0.519}, {0.522}} &{ {-0.373}, {0.454}, {0.057}}\\
    \hline


    \end{tabular}
\end{table*}

\begin{figure*}
\centering
\subfigure[Targeted Attack by KL loss]{\label{fig:edge-a}\includegraphics[height=0.18\linewidth]{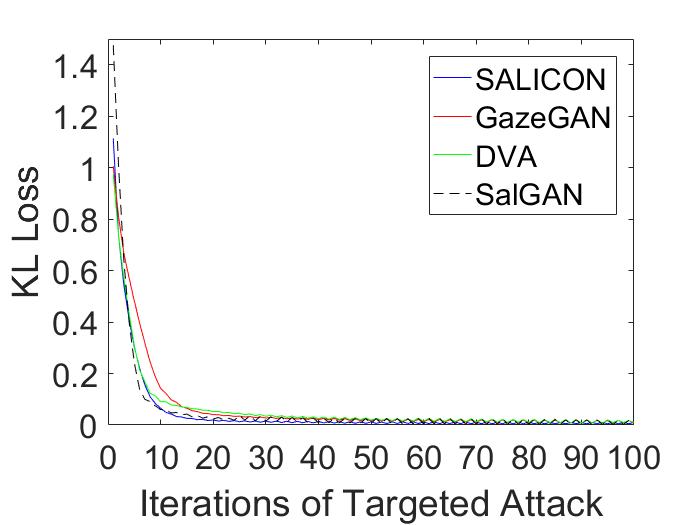}}
\subfigure[Targeted Attack by CC loss]{\label{fig:edge-a}\includegraphics[height=0.18\linewidth]{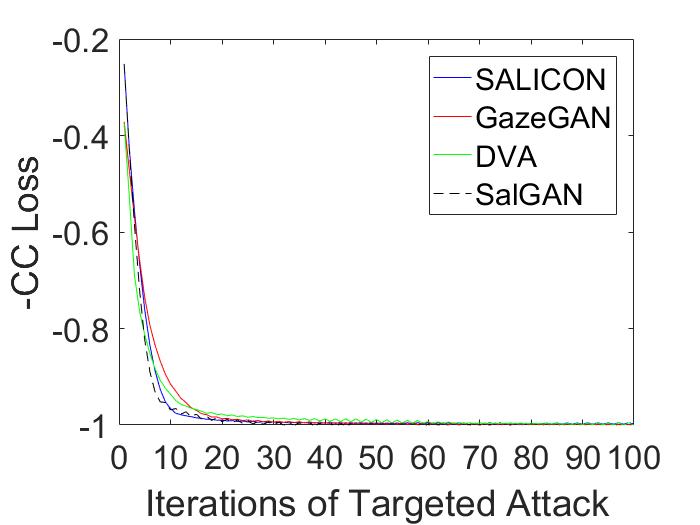}} 
\subfigure[Targeted Attack by NSS loss]{\label{fig:edge-a}\includegraphics[height=0.18\linewidth]{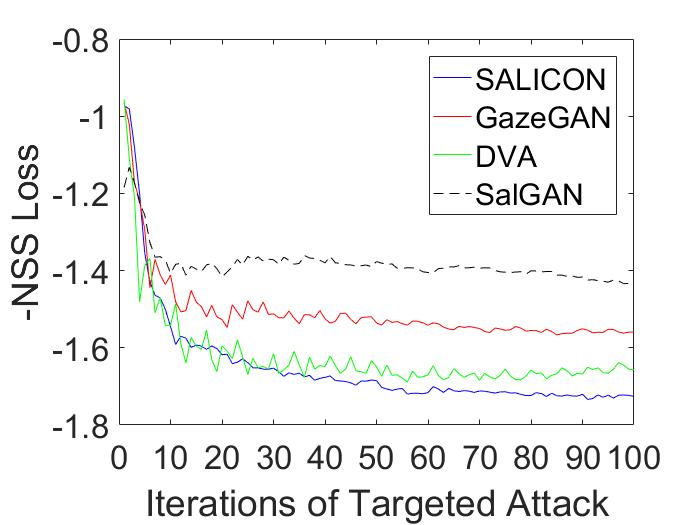}}
\subfigure[Targeted Attack by L$_1$ loss]{\label{fig:edge-a}\includegraphics[height=0.18\linewidth]{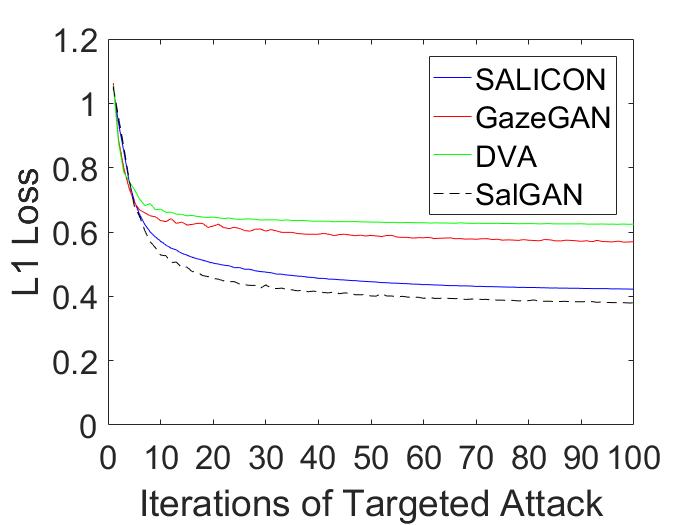}}\\

\subfigure[Nontargeted Attack by KL loss]{\label{fig:edge-a}\includegraphics[height=0.18\linewidth]{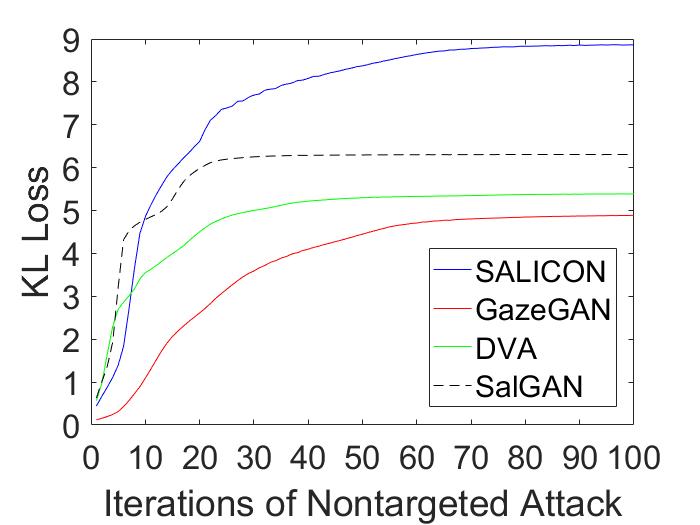}}
\subfigure[Nontargeted Attack by CC loss]{\label{fig:edge-a}\includegraphics[height=0.18\linewidth]{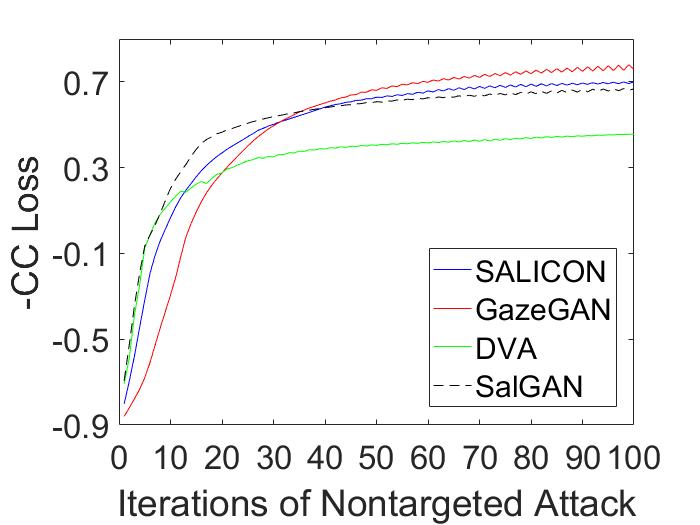}} 
\subfigure[Nontargeted Attack by NSS loss]{\label{fig:edge-a}\includegraphics[height=0.18\linewidth]{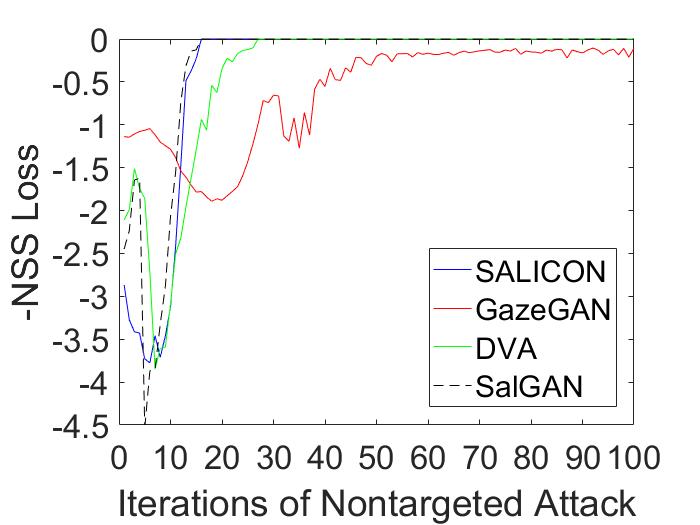}}
\subfigure[Nontargeted Attack by L$_1$ loss]{\label{fig:edge-a}\includegraphics[height=0.18\linewidth]{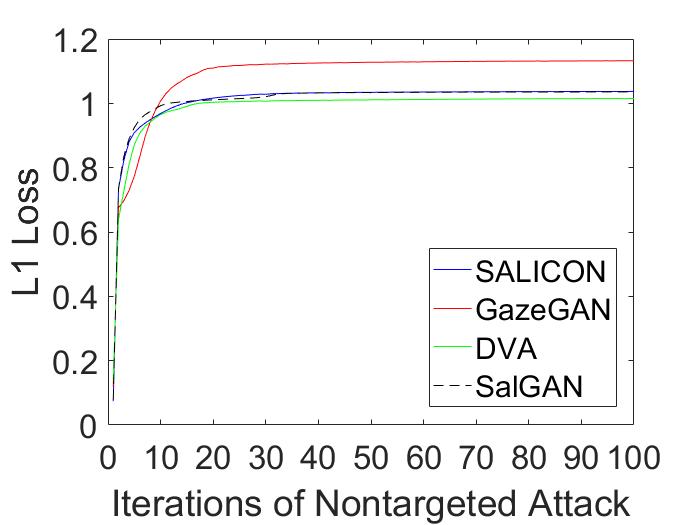}}

\caption{\small The convergence of the proposed attack when using different losses. }
\label{losses}
\end{figure*}

\subsection{Threat Model and Attack Performance}
As shown in Table \ref{tab:model}, we adopt 5 state-of-the-art deep networks and their variants to test the proposed method. These thread models are based on different base networks, and have different architectures. All of the threat models are trained and tested on the SALICON dataset \cite{SALICONDB}. More details about the architectures and parameters of these threat models are provided in the Supplementary Material. 

Experiments indicate that the proposed targeted and nontargeted attacks cause the significant performance drops for different saliency models.
Hendrycks~\etal~\cite{robust1} pointed out that multi-scale architectures achieve better robustness by propagating features across different scales at each layer rather than slowly gaining a global representation of the input as in traditional neural networks. For verifying this point, we further extend some existing single-stream saliency models to the multi-scale architectures, \eg~SalGAN \cite{SalGAN}, GazeGAN \cite{gazeganweb}, and DVA \cite{dva}. Specifically, we duplicate the single-stream network as two sub-networks, \ie~a coarse sub-network and a fine sub-network. Then we feed the original image and the downsampled image into the fine and coarse sub-networks, respectively. Finally, we concatenate the feature maps from coarse sub-network with the feature maps from fine sub-network in channel direction to predict final saliency map. However, experiments in Table \ref{tab:model} indicate that the multi-scale architectures are also highly vulnerable to adversarial attacks on saliency detection task.

\subsection{Convergence and Losses}
The convergences of proposed attack when using different losses are shown in Fig.\ref{losses}. For targeted attack, the losses measure the similarity between high-dimensional representations of guide image and the adversarial example, where the smaller losses mean the better similarity. We notice that, all of the losses decrease rapidly in the first 20 iterations, then go smoothly in the last iterations. KL and CC converge to the theoretical boundary values, \ie~KL=0 and CC=1. For nontargeted attack, the losses measure the similarity between the representations of original image and the adversarial example. KL, CC and L$_1$ losses increase rapidly in the first 20 iterations, while NSS loss decreases first, then increases rapidly and approximates the boundary value, \ie~NSS=0. This is because NSS metric calculates mean value of the normalized saliency map at fixation locations, and is sensitive to the false positives \cite{Bylinskii}. In  beginning iterations, the model attention on some secondary salient regions is distracted to the most salient regions, thus the mean value at obvious fixation locations is increased, \ie~-NSS value decreases. In later iterations, the model attention is further distracted away from the original salient regions until there is no overlapping regions with original fixation locations, \ie~NSS=0. We provide more visualizations about the convergence process in the Supplementary Material.

\subsection{Perceptibility, Agressivity, and Attacked Layer}
In this section, we explore the finer-grained correlation between the properties of attacked layer (\ie~layer depth and receptive field), and the quality (\ie~perceptibility and agressivity) of perturbations generated by this layer.

\begin{figure}
\centering
\subfigure[GazeGAN$_1$ Model]{\label{fig:edge-a}\includegraphics[height=0.29\linewidth]{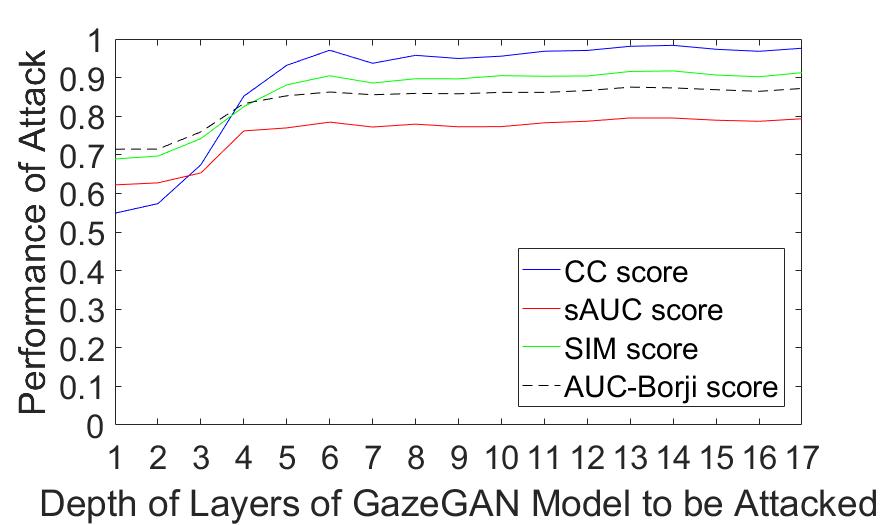}} 
\subfigure[SALICON$_1$ Model]{\label{fig:edge-a}\includegraphics[height=0.29\linewidth]{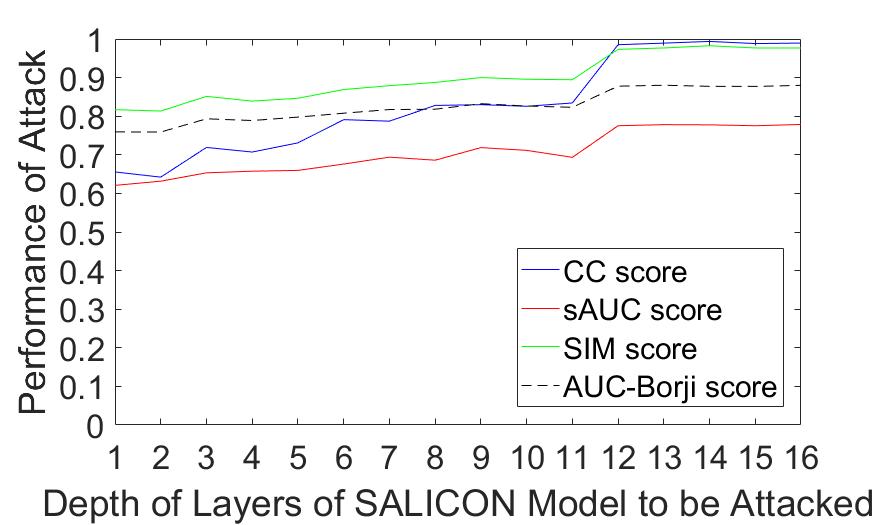}} 
\vspace{-5pt}
\caption{\small The relationship between the targeted attack performance and the depth of the attacked layer.}
\label{diflayer}
\end{figure}

We first investigate the relationship between the targeted attack performance and the depth of attacked layer, as shown in Fig.\ref{diflayer}. We adopt CC, sAUC, SIM and AUC-Borji metrics to measure the similarity between the predicted results of adversarial example and guide image.
Thus, the higher CC, sAUC, SIM and AUC-Borji scores represent the better targeted attack performance.
We notice that, in general, the deeper hidden layers achieve the better attack performance compared to the shallower layers. Besides, the attack performance will be greatly promoted at some intermediate hidden layers. For SALICON$_1$ model, the $12_{th}$ layer significantly increase the attack performance, because SALICON$_1$ adopts a coarse-scale and a fine-scale sub-networks, and the $12_{th}$ layer concatenates the feature maps from two sub-networks. This indicates that, for multi-scale architecture, attacking the layers of front-end sub-networks has slight impact on final model prediction, while attacking the deeper layers that pooling the features from sub-networks will revolutionize the final output.
For GazeGAN$_1$, the $6_{th}$ layer achieves a satisfying attack performance compared to the deeper layers including the image-space attack (\ie~the $17_{th}$ layer). In other word, if we obtain the information of the $1_{th}$-$6_{th}$ layers of GazeGAN$_1$, we can change the output of entire model completely.

\begin{figure}
\centering
\subfigure[GazeGAN$_1$ Model]{\label{fig:edge-a}\includegraphics[height=0.37\linewidth]{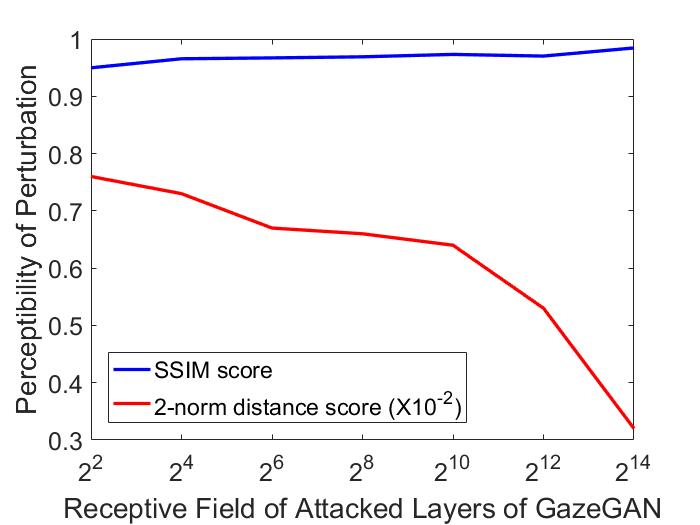}} 
\subfigure[SALICON$_1$ Model]{\label{fig:edge-a}\includegraphics[height=0.37\linewidth]{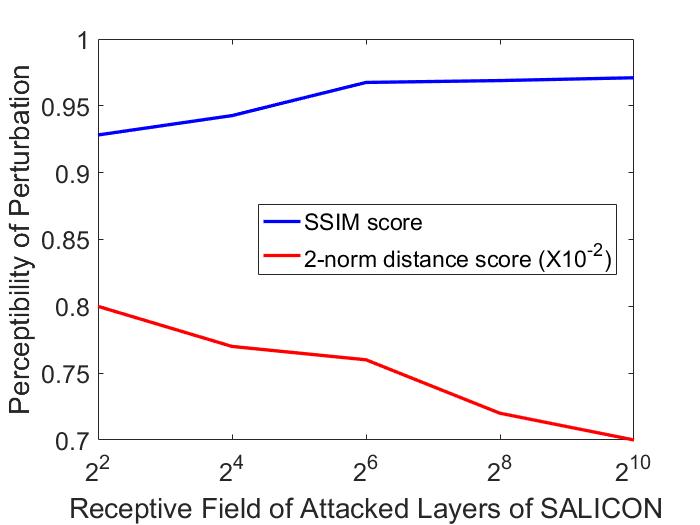}} 
\vspace{-5pt}
\caption{\small The relationship between perceptibility of the adversarial perturbation and receptive field of the attacked layer.}
\label{diflayer2}
\end{figure}

Next, we investigate the relationship between perceptibility of generated perturbation and the receptive field of attacked layer, as shown in Fig.\ref{diflayer2}. We use two metrics to evaluate the perceptibility of adversarial perturbation, \ie~SSIM and 2-norm distance. SSIM \cite{ssim} is a popular quality metric to measure the perceptual similarity of original image and adversarial example, and 2-norm distance \cite{tnnreview} measures the magnitude of perturbation. The higher SSIM and the lower 2-norm distance represent the lower perceptibility. Experiments indicate that the layers with the bigger receptive fields generate more imperceptible perturbations.

\begin{figure}
\centering
\subfigure[GazeGAN$_2$ Model]{\label{fig:edge-a}\includegraphics[height=0.36\linewidth]{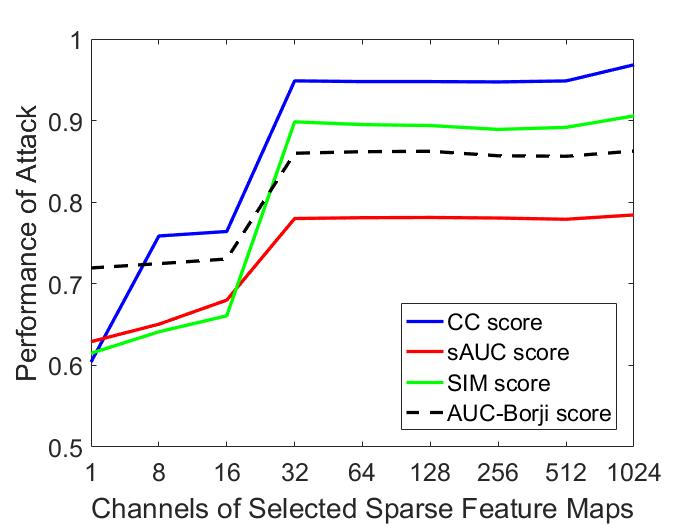}} 
\subfigure[SALICON$_1$ Model]{\label{fig:edge-a}\includegraphics[height=0.36\linewidth]{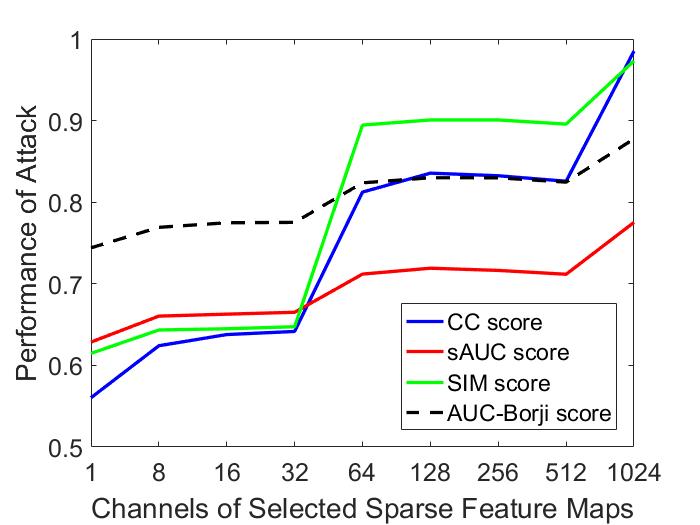}} 
\vspace{-5pt}
\caption{\footnotesize This figure explores how many sparse feature maps are able to perform a successful attack for a certain hidden layer.}
\label{diflayer3}
\end{figure}

\begin{figure}
\centering
\subfigure[ Feature-space Attack]{\label{fig:edge-a}\includegraphics[height=0.32\linewidth]{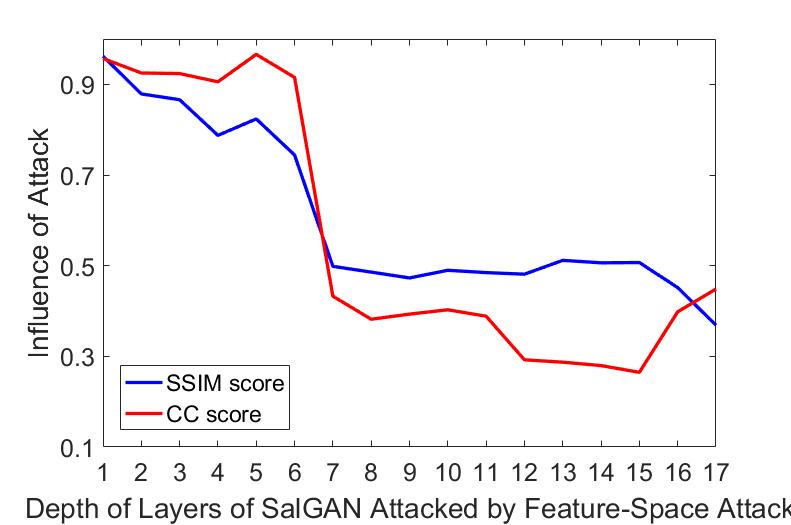}} 
\subfigure[ Image-space Attack]{\label{fig:edge-a}\includegraphics[height=0.32\linewidth]{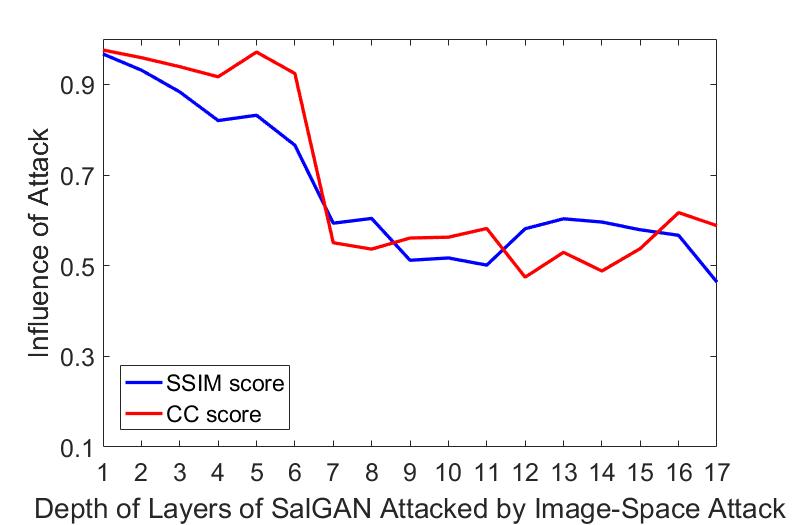}} 
\vspace{-5pt}
\caption{\small This figure explores where (at which layers) does the threat model (SalGAN$_1$) start to spoil when suffering from attacks.}
\label{diflayer4}
\end{figure}

Then, we explore that, for a certain hidden layer, how many sparse feature maps are qualified to perform a successful feature-space attack. The correlation between the attack performance and the amount of channels of sparse feature-maps is shown in Fig.\ref{diflayer3}. Specifically, we select 1, 8, 16 , 32, ... feature maps from the 1024 feature maps of the $10_{th}$ ($12_{th}$) hidden layer of GazeGAN$_2$ (SALICON$_1$) model, respectively. Then we compare the attack performance caused by different amount of feature channels. We notice that, for GazeGAN$_2$, 32 feature maps can generate a satisfying perturbation compared to using all 1024 feature maps. For SALICON$_1$, 64 feature maps can perform a satisfying attack, but there is still a gap compared to using all 1024 feature maps. In other word, GazeGAN$_2$ is easier to be attacked compared to SALICON$_1$, because there are too many similar and residual representations in feature space of GazeGAN$_2$. However, the performance of SALICON$_1$ on clean images is worse than GazeGAN$_2$.
We borrow the conclusion proposed by Su~\etal~\cite{eccvtradeoff} to explain this case: For image classification models, there is a clear trade-off between accuracy and robustness, and a better performance in testing accuracy reduces robustness. For saliency detection task, most of deep saliency models adopt similar base networks used in classification task as feature extractors.

Finally, we investigate where (at which layer) does the threat model start to spoil when suffering from feature-space and image-space attacks, as shown in Fig.\ref{diflayer4}. For each hidden layer, we adopt SSIM and CC metrics to measure the similarity between intermediate feature maps of original image and adversarial example.
The lower SSIM and CC values mean that the model attention has been distracted away from the salient regions of original image.
We unify the resolution (\ie~height$\times$width) of feature maps from different layers as 60$\times$80, for fair comparison.
We find that, SalGAN$_1$ model starts to spoil from the $7_{th}$ layer, which is a context hidden layer with a big receptive field, and extracts more high-level semantic information rather than low-level texture and edges. 
Besides, the feature-space and image-space attacks will destroy the SalGAN$_1$ model at a similar position, \ie~the $7_{th}$ hidden layer here.  

\subsection{Transferability and Countervailing Relation}
\begin{figure}
\centering
\subfigure[ Targeted Attack]{\label{fig:edge-a}\includegraphics[height=0.36\linewidth]{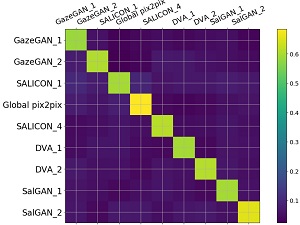}} 
\subfigure[ Nontargeted Attack]{\label{fig:edge-a}\includegraphics[height=0.36\linewidth]{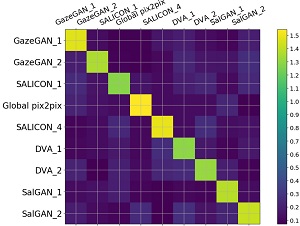}} 
\vspace{-5pt}
\caption{\footnotesize The transferability of proposed attack across different models. The vertical axis represents the target threat models, while the horizontal axis represents the source threat models. We use the perturbations generated by source models to attack the target models.}
\label{diflayer5}
\end{figure}

Fig.\ref{diflayer5} shows the transferability of proposed feature-space attack across different models. We adopt CC metric to measure the performance drop caused by different perturbations generated by different models. The higher performance drop represents a better transferability. Experiments indicate that transferability between different networks is weak, and the perturbations generated by other networks only result in a slight performance drop. This case is consistent with the attacks for segmentation and object detection tasks \cite{iccvsegadv}.
Besides, for single-stream and multi-scale models which utilize similar base networks, \ie~DVA$_1$ $vs$ DVA$_2$, SalGAN$_1$ $vs$ SalGAN$_2$, GazeGAN$_1$ $vs$ GazeGAN$_2$, the transferability is still weak. This indicates that the validity of proposed perturbation is highly depended on the thread model, including base network and architecture.

\begin{figure}
\centering
\subfigure[ Targeted Attack]{\label{fig:edge-a}\includegraphics[height=0.36\linewidth]{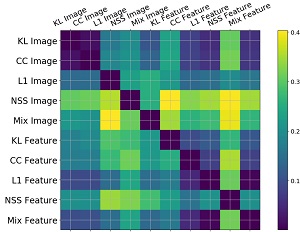}} 
\subfigure[ Nontargeted Attack]{\label{fig:edge-a}\includegraphics[height=0.36\linewidth]{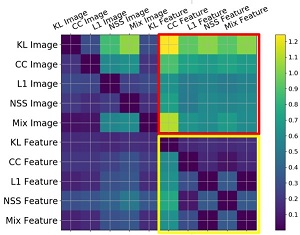}} 
\vspace{-5pt}
\caption{\footnotesize The countervailing relationship between different perturbations generated by different losses (Mix means the linear combination of KL, CC, NSS and L$_1$ losses), and generated by image-space and feature-space attacks. The vertical axis represents the target losses, while the horizontal axis represents the source losses. We use the perturbations generated by source losses to mitigate perturbations generated by target losses.}
\label{diflayer6}
\end{figure}

In Fig.\ref{diflayer6}, we explore weather the perturbations generated by different losses (or spaces) can mitigate each other. Specifically, we select one type of loss to generate a perturbation from image/feature space, and subtract it from the adversarial examples generated by unknown attacks. Then we adopt CC metric to measure the performance drop caused by the modified adversarial example. GazeGAN$_1$ serves as threat model here. Experiments indicate that, for targeted attack, perturbations generated by KL and CC losses from image-space can mitigate each other, but the other perturbations have discrepant patterns and are not able to countervail with each other. For nontargeted attack, the perturbations generated by KL and CC losses from image-space are able to mitigate other perturbations generated by different losses from both image and feature spaces to a certain extent. Besides, the feature-space perturbations can mitigate the perturbations generated by feature-space attack (\ie~regions in the yellow rectangle in Fig.\ref{diflayer6} (b)), but have slight impact on the perturbations from image-space (\ie~regions in the red rectangle in Fig.\ref{diflayer6} (b)). This indicates that the perturbations from feature-space have more similar patterns compared to perturbations from image-space. Besides, the perturbations from image-space are much denser, and can be harnessed to mitigate the sparser feature-space perturbations to a certain extent.

We perform a similar experiment as \cite{iccvsegadv}, which is randomly permuting the rows or columns of the adversarial perturbation. We find that the permuted perturbation fails to attack the models, indicating that the spatial structure of the perturbation is critical to the adversarial attack. However, we can not use this method to defend attack, because it also destroys the spatial information of original image.

\section{Conclusion}
We propose a sparse feature-space adversarial attack method against saliency  models for the first time. The proposed method generates a sparser and more visually imperceptible adversarial perturbation. Besides, it only requires partial information regarding the threat model, and completely changes the final prediction of the entire network. We further verify that the quality (\ie~aggressivity, perceptibility and diversity) of the generated adversarial perturbation is related to loss functions, the depth and receptive field of attacked layer,  and the amount of channels of sparse feature maps. Our work also provides lessons for devising defense method in the future: First, a good defense method  should consider the category of perturbations, because different losses and attacked layers will result in discrepant perturbation patterns. Second, defense method should discriminate the validity of perturbations in advance, because some invalid perturbations cannot disturb the model prediction at all, \eg~the perturbations generated by shallower hidden layers or generated by other model architectures.

{\small
\bibliographystyle{ieee}
\bibliography{egbib}

\begin{thebibliography}{10}

\bibitem{smautodrive}
S.~C. Yang and Y.~L. Hsu.
\newblock Full speed region sensorless drive of permanent-magnet machine
  combining saliency-based and back-emf-based drive.
\newblock {\em IEEE transactions on Industrial Electronics}, Vol. 64, No. 2,
  pp.1092-1101, 2017.

\bibitem{dataattmotodrive}
S.~Alletto, A.~Palazzi, F.~Solera, S.~Calderara, and R.~Cucchiara.
\newblock Dr (eye) ve: a dataset for attention-based tasks with applications to
  autonomous and assisted driving.
\newblock In {\em Proceedings of the IEEE Conference on Computer Vision and
  Pattern Recognition Workshops (pp. 54-60).}, pp. 54-60, 2016.

\bibitem{SalGAN}
J.~Pan, C.~Canton, K.~McGuinness, and et~al.
\newblock Salgan: Visual saliency prediction with generative adversarial
  networks.
\newblock In {\em arXiv preprint arXiv:1701.01081}, 2017.

\bibitem{SALICON}
X.~Huang, C.~Shen, X.~Boix, and Q.~Zhao.
\newblock Salicon: Reducing the semantic gap in saliency prediction by adapting
  deep neural networks.
\newblock In {\em Proceedings of IEEE International Conference on Computer
  Vision}, pp. 262-270, 2015.

\bibitem{MLnet}
M.~Cornia, L.~Baraldi, G.~Serra, and R.~Cucchiara.
\newblock A deep multi-level network for saliency prediction.
\newblock In {\em Proceedings of IEEE International Conference on Pattern
  Recognition}, pp. 3488-3493, 2016.

\bibitem{SalNet}
J.~Pan, K.~McGuiness, E.~Sayrol, N.~Conner, and et~al.
\newblock Shallow and deep convolutional networks for saliency prediction.
\newblock In {\em Proceedings of IEEE International Conference on Computer
  Vision and Pattern Recognization}, pp. 598-606, 2016.

\bibitem{SAM}
M.~Cornia, L.~Baraldi, G.~Serra, and et~al.
\newblock Predicting human eye fixations via an lstm-based saliency attentive
  model.
\newblock {\em IEEE Transactions on Image Processing}, Vol. 27, No. 10, pp.
  5142-5154, 2018.

\bibitem{borji2013state}
A.~Borji and L.~Itti.
\newblock State-of-the-art in visual attention modeling.
\newblock {\em IEEE Transactions on Pattern Analysis and Machine Intelligence},
  Vol. 35, No. 1, pp. 185-207, 2013.

\bibitem{borji2013quantitative}
A.~Borji, D.~N. Sihite, and L.~Itti.
\newblock Quantitative analysis of human-model agreement in visual saliency
  modeling: A comparative study.
\newblock {\em IEEE Transactions on Image Processing}, Vol. 22, No. 1, pp.
  55-69, 2013.

\bibitem{Borji_2013_ICCV}
A.~Borji, H.~R. Tavakoli, D.~N. Sihite, and L.~Itti.
\newblock Analysis of scores, datasets, and models in visual saliency
  prediction.
\newblock In {\em Proceedings of IEEE International Conference on Computer
  Vision}, pp. 921-928, 2013.

\bibitem{SalObjDec01}
A.~Oliva, A.~Torralba, M.~S. Castelhano, and J.~M. Henderson.
\newblock Top-down control of visual attention in object detection.
\newblock In {\em Proceedings of IEEE International Conference on Image
  Processing}, pp. I-253-I-257, 2003.

\bibitem{SalObjDec02}
S.~Frintrop.
\newblock A visual attention system for object detection and goal-directed
  search.
\newblock {\em Springer}, 2005.

\bibitem{SalSegment03}
A.~Mishra, Y.~Aloimonos, and C.~L. Fah.
\newblock Active segmentation with fixation.
\newblock In {\em Proceedings of IEEE 12th International Conference on Computer
  Vision}, pp. 468-475, 2009.

\bibitem{nipscompe}
A.~Kurakin, I.~Goodfellow, S.~Bengio, Y.~Dong, F.~Liao, M.~Liang, J.~Wang, and
  etc.
\newblock Adversarial attacks and defences competition.
\newblock In {\em In The NIPS'17 Competition: Building Intelligent Systems,
  Springer}, pp. 195-231, 2018.

\bibitem{fgsm}
I.~Goodfellow, J.~Shlens, and C.~Szegedy.
\newblock Explaining and harnessing adversarial examples.
\newblock In {\em Proceedings of the International Conference on Learning
  Representations}, 2015.

\bibitem{bfgs}
C.~Szegedy, W.~Zaremba, I.~Sutskever, J.~Bruna, D.~Erhan, I.~J. Goodfellow, and
  R.~Fergus.
\newblock Intriguing properties of neural networks.
\newblock In {\em Proceedings of the International Conference on Learning
  Representations}, 2014.

\bibitem{itefgsm}
A.~Kurakin, I.~Goodfellow, and S.~Bengio.
\newblock Adversarial examples in the physical world.
\newblock In {\em Proceedings of the Workshop of International Conference on
  Learning Representations}, 2016.

\bibitem{CandW}
N.~Carlini and D.~Wagner.
\newblock Towards evaluating the robustness of neural networks.
\newblock In {\em Proceedings of the IEEE Symposium on Security and Privacy},
  2017.

\bibitem{ATNnetworl}
S.~Baluja and I.~Fischer.
\newblock Adversarial transformation networks.
\newblock 2017.

\bibitem{iccvsegadv}
C.~Xie, J.~Wang, Z.~Zhang, Y.~Zhou, and L.~Xie an~A.~Yuille.
\newblock Adversarial examples for semantic segmentation and object detection.
\newblock In {\em Proceedings of the IEEE International Conference on Computer
  Vision}, pp. 1369-1378, 2017.

\bibitem{iccvuniper}
J.~H. Metzen, M.~C. Kumar, T.~Brox, and V.~Fischer.
\newblock Universal adversarial perturbations against semantic image
  segmentation.
\newblock In {\em Proceedings of the IEEE International Conference on Computer
  Vision}, pp. 2774-2783, 2017.

\bibitem{cvpr3d2d}
X.~Zeng, C.~Liu, Y.~Wang, W.~Qiu, L.~Xie, Y.~Tai, and A.~Yuille.
\newblock Adversarial attacks beyond the image space.
\newblock In {\em Proceedings of the IEEE International Conference on Computer
  Vision and Pattern Recognition}, 2019.

\bibitem{tnnreview}
X.~Yuan, P.~He, Q.~Zhu, and X.~Li.
\newblock Adversarial examples: Attacks and defenses for deep learning.
\newblock {\em IEEE transactions on neural networks and learning systems},
  2019.

\bibitem{resnetpaper}
K.~He, X.~Zhang, S.~Ren, and J.~Sun.
\newblock Deep residual learning for image recognition.
\newblock In {\em Proceedings of the IEEE International Conference on Computer
  Vision and Pattern Recognition}, pp. 770-778, 2016.

\bibitem{featureiclr}
S.~Sabour, Y.~Cao, F.~Faghri, and D.~Fleet.
\newblock Adversarial manipulation of deep representations.
\newblock In {\em Proceedings of the International Conference on Learning
  Representations}, 2016.

\bibitem{mit-saliency-benchmark}
Z.~Bylinskii, T.~Judd, A.~Borji, L.~Itti, F.~Durand, A.~Oliva, and A.~Torralba.
\newblock Mit saliency benchmark.
\newblock http://saliency.mit.edu/.

\bibitem{nssdef}
R.~Peters, A.~Iyer, L.~Itti, and C.~Koch.
\newblock Components of bottom up gaze allocation in natural images.
\newblock {\em Visual Research}, Vol. 45, No. 8, pp. 2397¨C2416, 2005.

\bibitem{Bylinskii}
Z.~Bylinskii, T.~Judd, A.~Oliva, A.~Torralba, and F.~Durand.
\newblock What do different evaluation metrics tell us about saliency models?
\newblock In {\em arXiv preprint cs.CV}, 2017.

\bibitem{gazeganweb}
Z.~Che, A.~Borji, G.~Zhai, G.~Guo, and P.L. Callet.
\newblock Gazegan: Invariance analysis and a robust new model.
\newblock {\em https://github.com/CZHQuality/Sal-CFS-GAN}, 2019.

\bibitem{pix2pixHD}
T.~C. Wang, M.~Y. Liu, J.~Y. Zhu, A.~Tao, J.~Kautz, and B.~Catanzaro.
\newblock High-resolution image synthesis and semantic manipulation with
  conditional gans.
\newblock In {\em Proceedings of the IEEE Conference on Computer Vision and
  Pattern Recognition}, pp. 8798-8807, 2018.

\bibitem{dva}
W.~Wang and J.~Shen.
\newblock Deep visual attention prediction.
\newblock {\em IEEE Transactions on Image Processing}, Vol.27, No.6, pp.
  2368-2378, 2018.

\bibitem{SALICONDB}
M.~Jiang, S.~Huang, J.~Duan, and Q.~Zhao.
\newblock Salicon: Saliency in context.
\newblock In {\em Proceedings of IEEE International Conference on Computer
  Vision and Pattern Recognition}, pp. 1072-1080, 2015.

\bibitem{robust1}
D.~Hendrycks and T.~G. Dietterich.
\newblock Benchmarking neural network robustness to common corruptions and
  surface variations.
\newblock In {\em arXiv preprint arXiv:1807.01697}, 2018.

\bibitem{ssim}
Z.~Wang, A.~C. Bovik, H.~R. Sheikh, and E.~P. Simoncelli.
\newblock Image quality assessment: from error visibility to structural
  similarity.
\newblock {\em IEEE Transactions on Image Processing}, Vol. 13, No. 4,
  pp.600-612, 2004.

\bibitem{eccvtradeoff}
D.~Su, H.~Zhang, H.~Chen, J.~Yi, P.Y. Chen, and Y.~Gao.
\newblock Is robustness the cost of accuracy?--a comprehensive study on the
  robustness of 18 deep image classification models.
\newblock In {\em Proceedings of the European Conference on Computer Vision},
  pp. 631-648, 2018.

\end{thebibliography}
}

\end{document}